\documentclass[letterpaper]{article}

\usepackage[preprint]{aaai2027}

\usepackage[hyphens]{url}
\usepackage{graphicx}
\usepackage{natbib}
\usepackage{caption}
\usepackage{amsmath}
\usepackage{amssymb}

\usepackage{booktabs}
\usepackage{multirow}
\usepackage{subfigure}

\usepackage{algorithm}
\usepackage{algorithmic}

\usepackage{newfloat}
\usepackage{listings}

\DeclareCaptionStyle{ruled}{
  labelfont=normalfont,
  labelsep=colon,
  strut=off
}

\floatstyle{ruled}
\newfloat{listing}{tb}{lst}{}
\floatname{listing}{Listing}

\title{Deliberate Before You Fly: Vision-Guided Spatial Deliberation for UAV See-and-Reach Navigation}

\author{
Fanfu Xue\textsuperscript{\rm 1},
En Yu\textsuperscript{\rm 2},
Bohang Liu\textsuperscript{\rm 1},
Hongjun Wang\textsuperscript{\rm 1},
Yang Yang\textsuperscript{\rm 1},
Xindi Wang\textsuperscript{\rm 3},
Jiande Sun\textsuperscript{\rm 4}
}

\affiliations{
\textsuperscript{\rm 1}School of Information Science and Engineering, Shandong University
\\
\textsuperscript{\rm 2}Faculty of Engineering and Information Technology,
University of Technology Sydney
\\
\textsuperscript{\rm 3}School of Artificial Intelligence, Shandong University
\\
\textsuperscript{\rm 4}School of Computer Science and Artificial Intelligence,
Shandong Normal University\\
    \{fanfuxue, liubohang\}@mail.sdu.edu.cn;
    isenn.yu@gmail.com; 
    \{hjw, yyang, xindi.wang\}@sdu.edu.cn;
    jiandesun@hotmail.com
}

\begin{document}

\maketitle

\begin{abstract}
UAV see-and-reach navigation requires an aerial agent to approach a language-specified target visible in its initial view and stop reliably near it. Existing methods typically map vision-language representations directly to action outputs without explicitly modeling intermediate fine-grained spatial decisions. This direct mapping causes semantic-control misalignment, leading to inconsistent maneuvers and unreliable termination. To address this issue, we propose DBFly, a vision-language waypoint prediction framework that introduces explicit vision-guided spatial deliberation before waypoint generation. Specifically, DBFly introduces a spatial maneuver decision chain that progressively performs target-direction anchoring, spatial diagnosis, and maneuver decision, enabling high-level maneuver intent to explicitly guide continuous waypoint generation. DBFly further constructs an implicit flight corridor by transforming the initial target-direction prior into a persistent geometric reference and deriving an online corridor state from the UAV's current position, thereby providing soft geometric guidance for spatial diagnosis and maneuver correction. In addition, DBFly develops a terminal-convergence-aware stopping strategy that characterizes terminal states through both target proximity and short-horizon motion convergence, enabling more reliable stopping near the target. Extensive experiments across seen, unseen-object, and unseen-scene test sets demonstrate that DBFly improves the success rate over the SOTA baseline by an average of 25.07 percentage points. The project homepage is available at https://xuefanfu.github.io/DBFly-Page.
\end{abstract}

\section{Introduction}
Aerial vision-and-language navigation (UAV-VLN) requires an agent to comprehend human instructions, perceive open-world environments, and execute autonomous flight~\cite{sun2026autofly,liu2024navagent}. Beyond long-range target search, successful UAV-VLN critically hinges upon the final reaching stage, where the agent must accurately ground a visually accessible target and translate language and egocentric observations into precise 3D motion. UAV \emph{see-and-reach} navigation specifically isolates this stage by assuming that the language-specified target is visible within the UAV’s initial field of view and requiring the agent to approach it through closed-loop waypoint prediction~\cite{xue2026see}. This setting centers on two closely coupled challenges: \emph{how to continuously anchor the target and perform appropriate target-oriented maneuvers}, and \emph{when to reliably terminate the flight near the target}.

\begin{figure}[]
\centering
\subfigure[Direct mapping]{
    \includegraphics[width=0.46\linewidth]{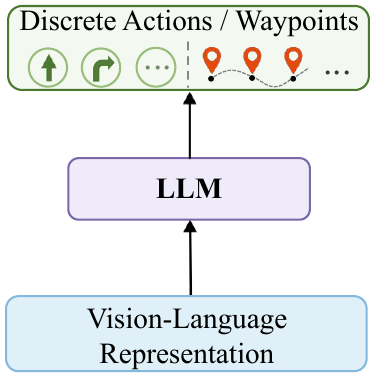}
    \label{fig:fig11}
    }
\subfigure[Spatial deliberation]{
    \includegraphics[width=0.46\linewidth]{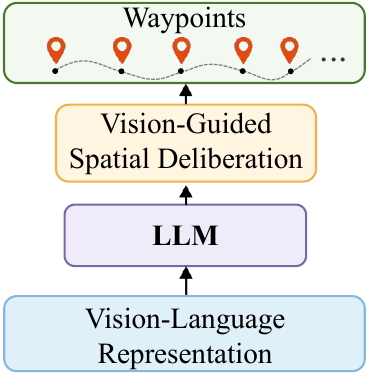}
    \label{fig:fig12}
    }
\caption{Comparison of two UAV-VLN navigation paradigms. (a) Existing methods directly map vision-language representations to action outputs, (b) whereas DBFly introduces vision-guided spatial deliberation before waypoint generation, as detailed in Fig.~\ref{DBFly_Framework}.}
\label{difference_pattern}
\end{figure}

Existing UAV-VLN methods generally adopt a direct vision-language-to-action mapping paradigm, as illustrated in Fig.~\ref{fig:fig11}. For example, UAV-ON~\cite{xiao2025uav} directly maps vision-language representations to discrete actions including stopping. In contrast, TravelUAV~\cite{wang2024towards}, SpatialFly~\cite{jiang2026spatialfly} and 3DG-VLN~\cite{xue2026see} directly map vision-language representations to continuous waypoints while relying on heuristic external detectors for flight termination. Despite their different designs, these methods bypass intermediate spatial cognition, producing maneuver and stopping signals without explicitly reasoning about how the UAV should maneuver or when it should terminate. Similar limitations have also drawn attention in ground-based VLN, where recent study demonstrates that explicitly modeling intermediate reasoning outperforms direct action prediction~\cite{10938647}. However, how to extend this reasoning paradigm to continuous 3D maneuvering and reliable termination in UAV see-and-reach navigation remains largely underexplored.

The direct mapping paradigm gives rise to two major limitations. \textbf{First}, bypassing explicit fine-grained spatial representations forces the model to directly bridge abstract vision-language concepts and low-level continuous control~\cite{wang2026cotfly,wu2025aeroduo,jiang2026dynfly}. This unbridled cross-domain translation easily causes semantic-control misalignment, where the generated waypoints deviate from the UAV’s actual target-relative spatial state, resulting in erratic maneuver execution. \textbf{Second}, directly relying on observations within the success radius to implicitly learn stopping decisions~\cite{xu2026aerialvla} lacks explicit evidence to determine the current execution phase and whether termination is warranted. Consequently, numerous intermediate states near the boundary dilute the distinctive supervisory cues of genuine expert-demonstrated terminations, severely complicating the learning of a reliable stopping boundary and often triggering premature or unstable flight termination.

To address these limitations, we propose \textbf{DBFly}, a vision-language waypoint prediction framework that introduces a novel \textbf{vision-guided spatial deliberation} mechanism before waypoint generation, as illustrated in Fig.~\ref{fig:fig12}. 
This deliberation explicitly answers two questions: \emph{how the UAV should maneuver} and \emph{when the flight should terminate}. It consists of a \textbf{spatial maneuver decision chain} and a \textbf{terminal-convergence-aware stopping strategy}. Specifically, spatial maneuver decision chain progressively performs target-direction anchoring, spatial diagnosis, and maneuver decision, enabling the model to infer control-relevant spatial decisions before waypoint generation. The resulting high-level maneuver intention explicitly modulates continuous waypoint generation, thereby improving the consistency between semantic understanding and executable flight control. Furthermore, DBFly constructs an implicit flight corridor by integrating the initial direction prior with the online corridor state, thereby providing soft geometric guidance for spatial diagnosis and maneuver correction. In addition, terminal-convergence-aware stopping strategy formulates stopping as a termination event jointly characterized by target proximity and short-horizon motion convergence, allowing the UAV to progressively approach the target, stabilize its motion, and terminate reliably in the target vicinity. 
The main contributions can be summarized as follows:
\begin{itemize}
    \item We propose \textbf{DBFly}, a novel UAV-VLN framework that pioneers vision-guided spatial deliberation before continuous waypoint generation. DBFly explicitly models intermediate fine-grained spatial decisions for maneuvering and termination, thereby enabling more consistent maneuvers and reliable termination.

    \item We design a spatial maneuver decision chain backed by an implicit flight corridor, which progressively transforms coarse priors into geometrically consistent maneuver intents to guide precise trajectory synthesis.
    
    \item We construct a terminal-convergence-aware stopping strategy that formulates stopping as a termination event, jointly considering target proximity and short-horizon motion convergence to achieve reliable flight termination.

    \item Extensive experiments demonstrate that DBFly consistently outperforms SOTA baselines, and real-world deployments further verify its robustness and effectiveness in practical see-and-reach navigation scenarios.
\end{itemize}

\section{Related Work}
Over the past two years, UAV-VLN has made initial progress in task formulation, method design, and dataset construction~\cite{zheng2026think,wang2026uav,chen2026aerialvla,zhou2026memory,gao2025openfly,zhang2025grounded,zheng2026onfly,wu2025vla,zhang2026apex,liu2026imagineuav,zheng2026worldfly,xu2026vgas}. Existing studies can be roughly grouped into top-down 2D-view navigation and egocentric 3D closed-loop navigation. The former performs language-conditioned route generation, target search, or mission planning on satellite imagery and city-scale geospatial maps, whereas the latter relies on onboard egocentric observations for closed-loop perception, decision-making, and motion control in 3D environments.

\paragraph{Top-Down 2D-View Navigation.}
\label{sec:top_down}
Built upon city-scale geospatial maps~\cite{hu2022sensaturban}, CityNav~\cite{lee2024citynav} introduces a language-goal aerial navigation dataset with geographic information. Following this setting, GeoNav~\cite{xu2025geonav} performs coarse-to-fine geospatial reasoning with structured spatial memory. HETT~\cite{ding2026history} adopts history-enhanced coarse-to-fine navigation, while HTNav~\cite{fan2026htnav} integrates imitation learning and reinforcement learning within a hybrid framework and enables tiered collaboration between macro-level path planning and fine-grained action control. In satellite-imagery-based scenarios, UAV-VLA~\cite{sautenkov2025uav} integrates satellite-image understanding with language reasoning for large-scale aerial mission generation, while UAV-VLPA~\cite{sautenkov2025uavvlpa} further performs global waypoint ordering via the traveling salesman problem and obstacle-aware local path refinement by A* algorithm~\cite{foead2021systematic}. Aerial vision-and-dialog navigation further extends this paradigm to interactive language-guided navigation~\cite{fan2023aerial,su2025learning,yu2026generalized,qi2026parse}.
These methods are effective for global landmark reasoning and large-scale route planning from top-down scenes. However, their 2D planning formulation largely ignores UAV-specific flight characteristics, such as egocentric visual feedback, altitude-aware 3D maneuvering, and closed-loop approach-and-stop control.

\begin{figure*}[th]
\centering
\includegraphics[width=\textwidth]{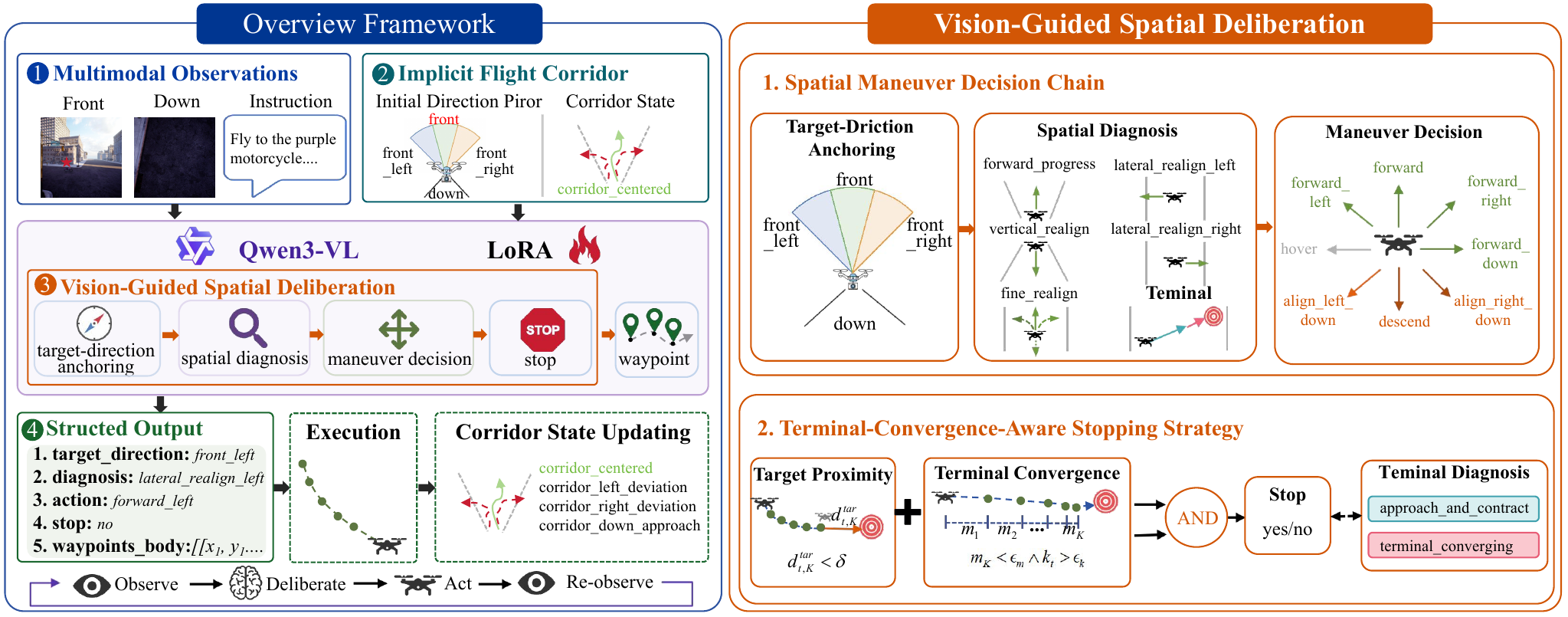}
\caption{\textbf{Overall framework of DBFly.} Built upon Qwen3-VL, DBFly integrates language instructions, egocentric visual observations, an initial target-direction prior, and the online corridor state to perform vision-guided spatial deliberation. The spatial maneuver decision chain and terminal-convergence-aware stopping strategy jointly guide the generation of geometrically consistent waypoints and reliable stopping decisions.}
\label{DBFly_Framework}
\end{figure*}

\paragraph{Egocentric 3D Closed-Loop Navigation.}
AerialVLN~\cite{liu2023aerialvln} first formulates UAV-VLN in 3D aerial environments, where the agent predicts fixed discrete actions from egocentric observations and fine-grained instructions. Following this paradigm, STMR~\cite{gao2024aerial} enhances LLM-based action prediction by projecting instruction-related semantic masks into a metric top-down representation, while Grid-based View Selection~\cite{zhao2025aerial} formulates action prediction as grid-based view selection and jointly considers horizontal-vertical action coupling. CityNavAgent~\cite{zhang2025citynavagent} reduces long-horizon navigation complexity through hierarchical semantic planning and global topological memory, and LookasideVLN~\cite{ning2026lookasidevln} exploits directional cues in instructions to construct egocentric lookaside graphs for efficient spatial reasoning. 

However, fixed discrete actions cannot fully exploit the maneuverability of UAVs, limiting flexible and precise flight control. TravelUAV~\cite{wang2024towards} predicts continuous 3D waypoints from egocentric observations and high-level target-oriented instructions, enabling more flexible UAV motion generation. Subsequent works further advance aerial navigation toward continuous and geometry-aware motion generation. SpatialFly~\cite{jiang2026spatialfly} aligns 2D semantic tokens with 3D geometric tokens through geometry-guided representation alignment, OpenVLN~\cite{lin2025openvln} improves long-horizon navigation through reinforcement learning and value-reward-guided trajectory synthesis. 3DG-VLN~\cite{xue2026see} isolates see-and-reach stage from the holistic search-and-reach process and introduces dynamic 3D direction guidance to maintain spatial alignment during closed-loop flight.

\section{Task Formulation}
\label{sec:task_formulation}
Given the multimodal observation, we formulate \textit{see-and-reach} navigation as a geometrically constrained spatial deliberation problem that predicts fine-grained structured decisions and short-horizon 3D waypoints. Let $\mathbf{P}_t=(\mathbf{p}_t,\mathbf{q}_t)$ denote the UAV pose at timestep $t$, where $\mathbf{p}_t\in\mathbb{R}^{3}$ is its world-frame position and $\mathbf{q}_t\in\mathbb{S}^{3}$ is its orientation quaternion. The corresponding body-to-world rotation matrix is given by $\mathbf{R}_t=\mathbf{R}(\mathbf{q}_t)\in\mathrm{SO}(3)$.
Following 3DG-VLN~\cite{xue2026see}, the coarse 3D direction space is defined as $\Omega_{\mathcal{D}}=\{\texttt{front\_left},\texttt{front},\texttt{front\_right},\texttt{down}\}$.

\paragraph{\textit{Observation Space.}} At each timestep $t$, the agent receives a multimodal observation $\mathcal{X}_t=\left(\mathcal{I},\mathcal{V}_t,\mathcal{D}_0,\mathcal{C}_t\right)$, where $\mathcal{I}$ is the navigation instruction, $\mathcal{V}_t=\{V_t^{f},V_t^{d}\}$ denotes the current front-view and downward-view observations, $\mathcal{D}_0\in\Omega_{\mathcal D}$ denotes the target's relative direction in the UAV's initial body frame, and $\mathcal{C}_t$ is the current implicit flight-corridor state, as illustrated in Fig.~\ref{DBFly_Framework}.

\paragraph{\textit{Spatial Deliberation Space and Policy Flow.}} Instead of directly generating continuous waypoints, the agent predicts
\begin{equation}
\mathcal{Y}_t=\left(\mathcal{D}_t,\mathcal{Z}_t,\mathcal{A}_t,\mathcal{S}_t,\mathcal{W}_t\right),
\label{eq:structured_output}
\end{equation}
where $\mathcal{D}_t\in\Omega_{\mathcal D}$ denotes the target's current relative direction in the UAV body frame, $\mathcal{Z}_t$ denotes the spatial diagnosis of the dominant navigation requirement and $\mathcal{A}_t$ denotes the maneuver decision, as shown in Fig.~\ref{DBFly_Framework}. Moreover, $\mathcal{S}_t\in\{\texttt{no},\texttt{yes}\}$ denotes the termination decision, and $\mathcal{W}_t=\{\mathbf{w}_{t+k}\}_{k=1}^{K}$ denotes a short-horizon waypoint sequence starting from $\mathbf{p}_t$ and expressed in the current UAV body frame. Each waypoint $\mathbf{w}_{t+k}=(\Delta x_{t+k},\Delta y_{t+k},\Delta z_{t+k})$ denotes the forward, rightward, and downward displacements, respectively.

The navigation policy is formulated as follows:
\begin{equation}
\mathcal{Y}_t=\pi_{\theta}\left(\mathcal{X}_t\right),
\label{eq:navigation_policy}
\end{equation}
where $\theta$ denotes the learnable model parameters. 

\paragraph{\textit{Objective and Success Criterion.}}
Let $\mathbf{p}_{\mathrm{tar}}$ denote the target position and
$\mathbf{p}_{\mathrm{end}}$ the terminal UAV position. An executed
trajectory $\tau$ is considered successful if
\begin{equation}
S(\tau)=\mathbb{I}\left[
\left\|\mathbf{p}_{\mathrm{end}}-\mathbf{p}_{\mathrm{tar}}\right\|_2
\leq \delta
\right],
\label{eq:success}
\end{equation}
where $\delta=10\,\mathrm{m}$ is the success radius. The task objective is to learn a navigation policy $\pi_{\theta}$ that maximizes the expected success rate $\mathbb{E}_{\tau\sim\pi_{\theta}}[S(\tau)]$.

\section{Method}
\label{sec:method}

\subsection{Overview}
As shown in Fig.~\ref{DBFly_Framework}, DBFly is built upon Qwen3-VL and adapts the pretrained model to UAV see-and-reach navigation through LoRA fine-tuning~\cite{hu2022lora}. It enables the agent to perform vision-guided spatial deliberation before waypoint generation. This deliberation comprises two tightly coupled components: the \textbf{spatial maneuver decision chain} that determines how the UAV should move toward the target, and the \textbf{terminal-convergence-aware stopping strategy} that determines whether navigation should continue or terminate. The former progressively performs target-direction anchoring, spatial diagnosis, and maneuver decision to formulate a high-level navigation intent that guides the generation of geometrically consistent waypoints. The latter jointly evaluates target proximity and short-horizon motion convergence to distinguish continued approach from stable terminal convergence, thereby supporting reliable stopping decisions. In addition, DBFly introduces an implicit flight corridor to provide soft geometric guidance for spatial diagnosis and maneuver correction. The following sections detail how the implicit flight corridor is constructed and how supervision signals are formulated for each component of vision-guided spatial deliberation.

\subsection{Implicit Flight Corridor}
\label{sec:implicit_corridor}

The implicit flight corridor is initialized from the initial target-direction prior, which is transformed into a persistent target-oriented geometric reference. Based on the UAV's current position relative to this reference, the online corridor state is derived to support spatial diagnosis and maneuver correction.

Specifically, for a down-view initial target direction, a persistent vertical geometric reference is maintained, as vertical alignment becomes the primary navigation requirement. For a front-view initial target direction, the corresponding horizontal geometric reference is represented by the following azimuthal support interval:
\begin{equation}
\mathcal{B}(\mathcal{D}_0)=
\begin{cases}
[-\alpha,-\gamma), & \mathcal{D}_0=\texttt{front\_left},\\
[-\gamma,\gamma], & \mathcal{D}_0=\texttt{front},\\
(\gamma,\alpha], & \mathcal{D}_0=\texttt{front\_right},
\end{cases}
\label{eq:implicit_corridor}
\end{equation}
where $\alpha$ represents the outer angular bound of the front-view range, and $\gamma$ separates the central $\texttt{front}$ region from the lateral $\texttt{front\_left}$ and $\texttt{front\_right}$ regions.

Given the initial UAV pose, we express the displacement from the initial to the current position in the initial UAV body frame as $\widetilde{\mathbf{p}}_t=\mathbf{R}_0^{\top}(\mathbf{p}_t-\mathbf{p}_0)$ and compute its horizontal displacement norm and azimuth:
\begin{equation}
\rho_t=\sqrt{\widetilde{p}_{t,x}^{2}+\widetilde{p}_{t,y}^{2}},\quad \beta_t=\operatorname{arctan2}(\widetilde{p}_{t,y},\widetilde{p}_{t,x}),
\label{eq:corridor_position}
\end{equation}
where $\rho_t$ and $\beta_t$ denote the horizontal displacement norm and azimuth, respectively.

The corridor state is determined from the geometric reference and the current UAV's horizontal displacement and azimuth. For a down-view corridor, the corridor state remains \texttt{corridor\_down\_approach} throughout the episode. For a front-view corridor, the UAV is considered to remain within the corridor when $\rho_t<\epsilon_c$ or $\beta_t\in\mathcal{B}(\mathcal{D}_0)$, and is therefore assigned the \texttt{corridor\_centered} state. Otherwise, it is assigned to \texttt{corridor\_left\_deviation} when $\beta_t$ falls below the lower corridor bound and to \texttt{corridor\_right\_deviation} when it exceeds the upper bound.

\subsection{Spatial Maneuver Decision Chain}
\label{sec:spatial_deliberation}
The spatial maneuver decision chain consists of target-direction anchoring, spatial diagnosis, and maneuver decision. Specifically, target-direction anchoring determines the coarse target direction relative to the current UAV body frame, spatial diagnosis identifies the dominant navigation requirement at the current stage, and maneuver decision specifies the high-level motion intention to be executed. This progressive reasoning process provides explicit semantic guidance for subsequent waypoint generation.

\paragraph{\textit{Target-Direction Anchoring.}} The target-direction anchoring serves as the first stage of spatial deliberation, providing a coarse 3D directional cue $\mathcal{D}_t\in\Omega_{\mathcal{D}}$ for subsequent spatial diagnosis and maneuver decision.

\paragraph{\textit{Maneuver Decision.}}
The maneuver decision is constructed by abstracting the dominant geometric motion pattern from the expert short-horizon waypoints. Since future waypoints are relative to the current UAV position, we set $\mathbf{w}_t=\mathbf{0}$. The terminal horizontal azimuth and the mean incremental displacement are defined as
\begin{equation}
\begin{gathered}
\theta_t^{w}
=\operatorname{arctan2}
\left(\Delta y_{t+K},\Delta x_{t+K}\right),\\
\delta\mathbf{w}_{t+k}
=\mathbf{w}_{t+k}-\mathbf{w}_{t+k-1},\\
\boldsymbol{\mu}_t
=\frac{1}{K}\sum_{k=1}^{K}\delta\mathbf{w}_{t+k}
=(\mu_t^x,\mu_t^y,\mu_t^z),
\end{gathered}
\label{eq:trajectory_motion_features}
\end{equation}
where $\theta_t^{w}$ represents horizontal azimuth, while $\boldsymbol{\mu}_t$ represents the mean incremental displacement over the short horizon.

Specifically, for samples with $\mathcal{S}_t=\texttt{yes}$, the maneuver decision is set to \texttt{hover}, indicating terminal convergence and suppressing large maneuvers while allowing minor adjustments near the target. For the remaining samples, downward-dominant motion is identified when
\begin{equation}
\Delta z_{t+K}>\epsilon_d,
\quad
|\Delta x_{t+K}|<\epsilon_s,
\label{eq:downward_motion_identification}
\end{equation}
where $\epsilon_d$ and $\epsilon_s$ denote the minimum downward-displacement threshold and the maximum forward-displacement threshold, respectively. Within this mode, samples satisfying $\Delta y_{t+K}<-\epsilon_l$ and $\Delta y_{t+K}>\epsilon_l$ are assigned to \texttt{align\_left\_down} and \texttt{align\_right\_down}, respectively, while the remaining samples are assigned to \texttt{descend}. Here, $\epsilon_l$ denotes the lateral-displacement threshold for distinguishing lateral alignment from direct descent.

For samples exhibiting sufficient forward displacement,
\begin{equation}
\Delta x_{t+K}>\epsilon_f,
\label{eq:forward_motion_identification}
\end{equation}
where $\epsilon_f$ denotes the minimum terminal forward-displacement threshold, the terminal horizontal azimuth is used to determine the lateral maneuver mode. Specifically, samples with $\theta_t^{w}<-\gamma$ and $\theta_t^{w}>\gamma$ are assigned to \texttt{forward\_left} and \texttt{forward\_right}, respectively. For samples not assigned to either lateral forward mode, coupled forward-downward motion is identified when
\begin{equation}
\mu_t^x>\epsilon_{\mu x},
\quad
\mu_t^z>\epsilon_{\mu z},
\label{eq:forward_down_motion}
\end{equation}
where $\epsilon_{\mu x}$ and $\epsilon_{\mu z}$ denote the thresholds for the mean incremental displacements along the forward and downward directions, respectively, and samples satisfying both conditions are assigned to \texttt{forward\_down}. The remaining samples satisfying $\Delta x_{t+K}>\epsilon_f$ are assigned to \texttt{forward}.

\paragraph{\textit{Spatial Diagnosis.}}
The spatial diagnosis captures the dominant navigation requirement required under the current navigation context and provides a spatial requirement guidance for the subsequent maneuver decision. Its supervision is hierarchically constructed from the current corridor state, maneuver decision, target direction, and expert waypoint geometry.

For samples exhibiting a lateral corridor deviation, corridor recovery takes precedence over the action-induced diagnosis. When the UAV deviates to the left boundary of the corridor, maneuvers with a rightward motion component are assigned the diagnosis \texttt{lateral\_realign\_right}. A \texttt{forward} maneuver is assigned the same diagnosis when $\Delta y_{t+K}>-\epsilon_r$, where $\epsilon_r$ denotes the maximum tolerated lateral displacement opposite to the required recovery direction. Conversely, when the UAV deviates to the right boundary of the corridor, maneuvers with a leftward motion component are assigned the diagnosis \texttt{lateral\_realign\_left}, while a \texttt{forward} maneuver is assigned the same diagnosis when $\Delta y_{t+K}<\epsilon_r$. Samples whose maneuver direction is incompatible with the required corridor recovery are excluded from supervision construction.

When the UAV flight within the corridor, the diagnosis is determined primarily by the maneuver decision. Maneuvers with a leftward motion component are mapped to \texttt{lateral\_realign\_left}, whereas those with a rightward motion component are mapped to \texttt{lateral\_realign\_right}. Maneuvers involving downward motion are assigned to \texttt{vertical\_realign}. For a \texttt{forward} maneuver, the diagnosis is set to \texttt{forward\_progress} when the target lies in the \texttt{front} region and to \texttt{fine\_realign} otherwise, indicating forward-dominant motion that requires local alignment refinement. Terminal-related diagnoses are described together with the stopping strategy in the following section.

\subsection{Terminal-Convergence-Aware Stopping Strategy}
\label{sec:terminal_stopping}

\paragraph{\textit{Terminal-Convergence Evidence.}}
Reliable stopping should indicate stable terminal convergence rather than mere entry into the success radius. We therefore construct the terminal supervision from two complementary types of evidence: target proximity and short-horizon motion convergence. Let $d_{t,K}^{\mathrm{tar}}$ denote the distance between the position of the last expert waypoint and the target position. The evolution of short-horizon motion can be defined as
\begin{equation}
\begin{gathered}
m_{k}=\left\|\delta\mathbf{w}_{t+k}\right\|_2,
\quad k=1,\ldots,K,\\
\kappa_t=m_1-m_K,
\end{gathered}
\label{eq:trajectory_convergence}
\end{equation}
where $m_k$ denotes the displacement magnitude between adjacent waypoints, while $\kappa_t$ indicates a contracting motion trend. The proximity and convergence evidence are then defined as
\begin{equation}
\begin{gathered}
b_t^{\mathrm{prox}}
=\mathbb{I}\left[d_{t,K}^{\mathrm{tar}}<\delta\right],\\
b_t^{\mathrm{conv}}
=\mathbb{I}\left[m_K<\epsilon_m \,\wedge\, \kappa_t >\epsilon_{\kappa}\right],
\end{gathered}
\label{eq:terminal_evidence}
\end{equation}
where $b_t^{\mathrm{prox}}$ indicates whether the final expert waypoint lies within the success radius $\delta$, and $b_t^{\mathrm{conv}}$ indicates whether the terminal motion is sufficiently small and exhibits sufficient contraction. $\epsilon_m$ and $\epsilon_{\kappa}$ denote the terminal-motion and motion-contraction thresholds, respectively.

The stopping supervision is determined by its conjunction:
\begin{equation}
\mathcal{S}_t=
\begin{cases}
\texttt{yes}, & b_t^{\mathrm{prox}}b_t^{\mathrm{conv}}=1,\\
\texttt{no}, & \text{otherwise}.
\end{cases}
\label{eq:stop_supervision}
\end{equation}

\paragraph{\textit{Terminal-Related Diagnosis.}}
Before constructing any regular spatial diagnosis, we first establish the terminal-related diagnosis based on the terminal evidence. When $\mathcal{S}_t=\texttt{yes}$, the diagnosis is set to \texttt{terminal\_converging}. Otherwise, trajectories satisfying $b_t^{\mathrm{prox}}=1$ and $\kappa_t>\epsilon_{\kappa}$ are labeled as \texttt{approach\_and\_contract}, indicating that the UAV has entered the target region but has not yet converged.

\subsection{Structured Supervised Optimization}
\label{sec:optimization}
The spatial deliberation fields are not independent classification targets. Their ordering encodes the progression from spatial understanding to physical actuation. We therefore serialize the complete output $(\mathcal{D}_t,\mathcal{Z}_t,\mathcal{A}_t,\mathcal{S}_t,\mathcal{W}_t)$ and optimize it autoregressively, allowing later fields to condition on the previously generated reasoning decisions. For a training set containing $N$ samples, the supervised objective is
\begin{equation}
\mathcal{L}_{\mathrm{SFT}}=-\frac{1}{N}\sum_{n=1}^{N}\sum_{j=1}^{L_n}\log\pi_{\theta}\left(y_j^{(n)}\mid\mathcal{X}^{(n)},y_{<j}^{(n)}\right),
\label{eq:sft_objective}
\end{equation}
where $L_n$ is the length of the serialized target sequence. This objective jointly learns the output format, fine-grained spatial decisions and continuous waypoints without introducing separate prediction heads or auxiliary losses.

\begin{table*}
\centering
\begin{tabular}{ccccccccccccc} 
  \toprule
  \multirow{2}{*}{\textbf{Methods}} & \multicolumn{4}{c}{Test} & \multicolumn{4}{c}{Test UO}  &  \multicolumn{4}{c}{Test US}\\
  \cmidrule(lr){2-5}   \cmidrule(lr){6-9}   \cmidrule(lr){10-13} 
  &SR \textuparrow & OSR \textuparrow &NE \textdownarrow & SPL \textuparrow &SR \textuparrow & OSR \textuparrow &NE \textdownarrow & SPL \textuparrow &SR \textuparrow & OSR \textuparrow &NE \textdownarrow & SPL \textuparrow  \\
  \midrule
Random&  0.66 & 0.66 & 46.34 & 0.07 & 0.00 & 1.22 & 60.66 & 0.00 & 0.00 & 1.73 & 61.45 & 0.00 \\
Fixed& 0.00 & 3.29  & 46.30 & 0.00 & 0.00 & 2.44 & 60.51 & 0.00 & 0.00 & 0.00 & 59.54 & 0.00  \\
Qwen2.5-VL-7B & 0.00 & 3.29  & 42.31 & 0.00 & 0.00 & 3.05 & 56.53 & 0.00 & 0.58 & 2.31 & 55.36 & 0.04 \\
Qwen3-VL-8B & 0.66 & 28.95  & 58.44 & 0.28 & 1.22 & 14.63 & 56.44 & 0.47 & 0.58 & 24.86 & 55.21 & 0.08 \\
TravelUAV& 25.00   & 30.26 & 34.03  & 17.89& 20.12 & 24.39 & 37.30 & 16.08 & 16.76 & 24.28 & 38.75 & 12.15 \\
3DG-VLN & 38.82 & 55.92 & 24.41 & 27.70 & 28.05 & 39.63 & 35.07 & 21.92 & 21.39 & 35.84 & 37.73 &13.43 \\
  \midrule
DBFly & \textbf{62.50} & \textbf{71.05} & \textbf{16.83} & \textbf{42.37} & \textbf{51.83} & \textbf{63.41} & \textbf{21.39} & \textbf{36.07} & \textbf{49.13} & \textbf{57.23} & \textbf{22.21} & \textbf{31.24} \\
    \midrule
\end{tabular}
\caption{Comparison with baseline methods. \textbf{Bold} indicates the best result.}
\label{comparison_studies}
\end{table*}

\begin{figure*}
\centering
\includegraphics[width=\textwidth]{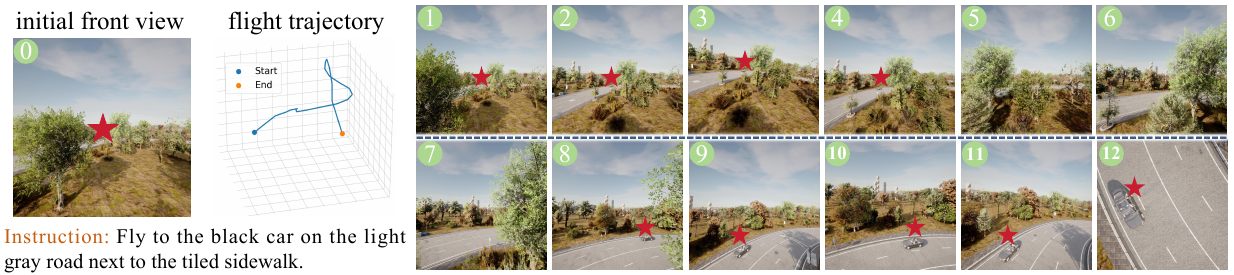}
\caption{Visualization of DBFly navigation in the high-fidelity simulation environment. The red star indicates the approximate target location, while its absence means that the target is outside the UAV's current field of view. }
\label{simulation_visual_1}
\end{figure*}

\section{Experiments}
\subsection{Experiment Settings}

\paragraph{Datasets.}
We evaluate DBFly on the UAV-VLN-FOV benchmark~\cite{xue2026see}, which contains 2,717 trajectories across 14 scenarios and 89 target categories. The dataset is divided into a training set with 2,228 trajectories and three test sets: the seen test set (Test), unseen-object test set (Test UO), and unseen-scene test set (Test US), containing 152, 164, and 173 trajectories, respectively. 

\paragraph{Baselines.} We compare DBFly with six baselines: \textit{Random}, \textit{Fixed}, \textit{Qwen2.5-VL-7B}~\cite{bai2025qwen25vltechnicalreport}, \textit{Qwen3-VL-8B}~\cite{bai2025qwen3}, \textit{TravelUAV}~\cite{wang2024towards}, and \textit{3DG-VLN}~\cite{xue2026see}. Among them, \textit{Random} and \textit{Fixed} serve as lower-bound references, illustrating the difficulty of see-and-reach navigation without intelligent guidance, following the settings in~\cite{xue2026see}. \textit{Qwen2.5-VL-7B} and \textit{Qwen3-VL-8B} are evaluated in a zero-shot setting to assess the capabilities of general-purpose vision-language models for 3D spatial reasoning and waypoint planning without task-specific adaptation. \textit{TravelUAV} and \textit{3DG-VLN} are SOTA task-specific UAV-VLN methods that directly map vision-language representations to 3D waypoints.

\paragraph{Metrics.}
We evaluate all methods using four standard metrics: Success Rate (SR), Oracle Success Rate (OSR), Navigation Error (NE), and Success-weighted Path Length (SPL).

\paragraph{Implementation Details.}
\label{sec:implementation_details}
We adopt Qwen3-VL-8B as the base model for DBFly to balance navigation capability and computational cost. All training experiments are conducted on 2 NVIDIA Tesla A100 40GB GPUs. DBFly is fine-tuned for 2 epochs using LoRA with a rank of 8, a batch size of 1, 16 gradient accumulation steps, a learning rate of $10^{-4}$, and a warmup ratio of 0.1. To balance the maneuver space of spatial deliberation, we downsample \texttt{forward} samples by 50\% and oversample any other category below 10\% of this retained count up to 10\%, leaving the rest unchanged, using a random seed of 42. Additional implementation details are provided in Appendix~\ref{app:experiment_details}.

\subsection{Comparisons with Baselines} 
\label{comparisons}
\paragraph{Quantitative Evaluation.} As shown in Table~\ref{comparison_studies}, DBFly consistently outperforms all baselines across all test sets. Specifically, compared with the strongest baseline, 3DG-VLN, DBFly improves SR and SPL on Test by 23.68 and 14.67 percentage points, respectively, demonstrating superior target-reaching capability and navigation efficiency. Under the more challenging Test UO and Test US datasets, these performance advantages remain equally pronounced, yielding SR improvements of 23.78 and 27.74 percentage points and SPL gains of 14.15 and 17.81 percentage points, respectively. The consistent improvements further confirm that introducing vision-guided spatial deliberation before waypoint generation improves semantic-control alignment, enabling the model to infer precise maneuver decisions and identify terminal states, thereby achieving superior maneuver consistency and termination reliability.

\begin{table*}
\centering
\begin{tabular}{ccccccccccccc} 
  \toprule
  \multirow{2}{*}{\textbf{Strategies}} & \multicolumn{4}{c}{Test} & \multicolumn{4}{c}{Test UO}  &  \multicolumn{4}{c}{Test US}\\
  \cmidrule(lr){2-5}   \cmidrule(lr){6-9}   \cmidrule(lr){10-13} 
  &SR \textuparrow & OSR \textuparrow &NE \textdownarrow & SPL \textuparrow &SR \textuparrow & OSR \textuparrow &NE \textdownarrow & SPL \textuparrow &SR \textuparrow & OSR \textuparrow &NE \textdownarrow & SPL \textuparrow  \\
  \midrule
w/o VGSD & 40.79 & 55.26 & 25.49 & 29.32 & 27.44 & 34.76 & 32.57 & 18.99 & 29.48 & 40.46 & 33.97 & 18.64 \\
w/o SMDC &  28.29 & 59.21 & 33.57 & 18.06 & 15.85 & 35.98 & 36.20 & 10.83 & 19.65 & 45.66 & 37.94 & 11.19 \\
w/o Corr & 58.55 & 67.11 & 18.65 & 38.93 & 48.17 & 56.71 & 26.29 & 35.76 & 47.40 & 53.76 & \textbf{22.07} & 30.48 \\
w/o Stop & 49.34 & 55.26 & 21.36 & 35.76& 40.24 & 46.34 & 28.59 & 29.15 & 33.53 & 37.57 & 26.18 & 22.03 \\
\midrule
DBFly & \textbf{62.50} & \textbf{71.05} & \textbf{16.83} & \textbf{42.37}  & \textbf{51.83} & \textbf{63.41} & \textbf{21.39} & \textbf{36.07} &\textbf{49.13} & \textbf{57.23} & 22.21 & \textbf{31.24} \\
    \midrule
\end{tabular}
\caption{Ablation study of different strategies. w/o denotes removal of the component. VGSD: vision-guided spatial deliberation; SMDC: spatial maneuver decision chain; Corr: implicit flight corridor; Stop: terminal-convergence-aware stopping.}
\label{ablation_studies}
\end{table*}

\begin{figure*}
\centering
\includegraphics[width=\textwidth]{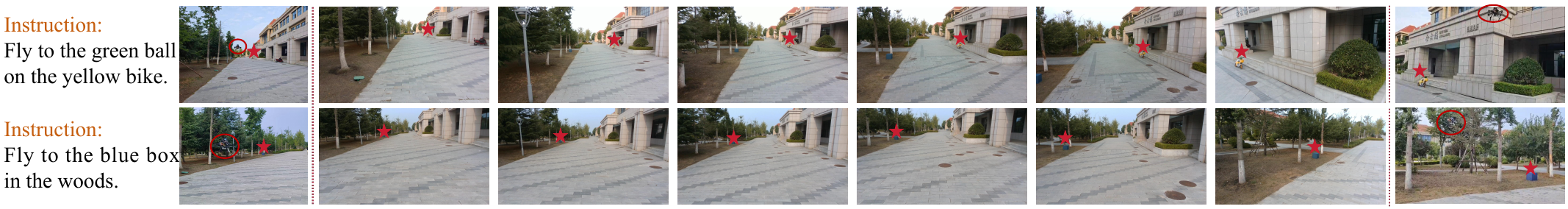}
\caption{Visualization of DBFly navigation in real-world environments. In each row, the first and last images show global views at the initial and terminal positions, respectively, while the intermediate images present representative onboard front-view observations collected during flight.}
\label{real_visual_1}
\end{figure*}

\paragraph{Qualitative Evaluation.} 
Fig.~\ref{simulation_visual_1} visualizes the observations and corresponding flight trajectory of an episode. At stages 5-7, although the target disappears from the UAV's field of view, DBFly progressively corrects its flight direction and reorients toward the target through autonomous spatial reasoning. Meanwhile, when encountering obstacles, DBFly plans a collision-free path around them, demonstrating effective obstacle-avoidance capability. These observations further indicate that vision-guided spatial deliberation endows DBFly with autonomous spatial reasoning, enabling it to diagnose evolving spatial conditions, determine appropriate maneuver, and translate them into reliable flight control. More qualitative results are provided in the Appendix~\ref{app:additional_experiments}.

\subsection{Ablation Study} 
\label{ablation}
Moreover, we conduct ablation studies to evaluate the effectiveness of the proposed components, as shown in Table~\ref{ablation_studies}. Removing VGSD decreases SR by 21.71, 24.39, and 19.65 percentage points on Test, Test UO, and Test US, respectively, demonstrating that VGSD substantially enhances the model's autonomous spatial understanding and decision-making capabilities, thereby improving both motion planning and termination reliability. Within VGSD, removing SMDC results in an average SR decrease of 33.22 percentage points across the three test sets, indicating its substantial contribution to navigation performance. The result highlights the importance of progressively transforming target-direction anchoring, spatial diagnosis and maneuver decision into high-level maneuver intentions that guide consistent waypoint generation. In addition, as geometric support for SMDC, the implicit flight corridor contributes an average SR gain of 3.11 percentage points, indicating that soft geometric constraints contribute to more stable spatial diagnosis and maneuver correction. Finally, removing terminal-convergence-aware stopping strategy leads to an average SR decrease of 13.45 percentage points, highlighting the importance of explicitly distinguishing continued target approach from stable terminal convergence for reliable termination. 

Notably, removing VGSD achieves a higher SR than removing SMDC, indicating a strong coupling between the stopping decision and the terminal-related spatial diagnoses. When these diagnoses are removed, the model lacks an explicit evidence for learning when to stop, leading to poorer stopping performance than the threshold-based stopping strategy. Furthermore, we perform ablation experiments on sampling to isolate its impact, confirming that the performance gains stem from the proposed deliberation supervision rather than from sampling alone, with detailed experimental results provided in Appendix~\ref{app:additional_experiments}.

\subsection{Real-World Experiments}
\label{real_world}
To assess its practical applicability, we further deploy DBFly in real-world flight tests. As shown in Fig.~\ref{real_visual_1}, DBFly continuously predicts precise 3D waypoints, enabling the UAV to fly toward the language-specified target. Two capabilities are particularly evident. First, during the terminal approach, DBFly continuously adjusts its maneuver according to the target-relative spatial state and actively flies toward the specified target. Second, by distinguishing continued target approach from stable terminal convergence, DBFly enables reliable stopping near the target. These results provide compelling evidence that the vision-guided spatial deliberation learned by DBFly effectively generalizes from simulation to real-world environments, supporting both precise target approach and reliable termination.

\section{Conclusion and Limitations}
\label{conclusion}
In this paper, we propose DBFly, a vision-language waypoint prediction framework that introduces vision-guided spatial deliberation for UAV see-and-reach navigation. To explicitly model intermediate spatial decisions, DBFly incorporates a spatial maneuver decision chain to enhance semantic-to-control consistency and a terminal-convergence-aware stopping strategy to boost stopping reliability. Furthermore, DBFly constructs an implicit flight corridor to provide soft geometric guidance for spatial diagnosis and maneuver correction. Extensive experiments demonstrate that DBFly achieves SOTA performance, while real-world flight tests further confirm both its operational reliability and its practical deployability.

Nevertheless, DBFly does not explicitly predict future visual states to assess action feasibility before execution. Future work will integrate spatial deliberation with an action-conditioned world model to predict future visual states, assess spatial feasibility before execution, and refine decisions via closed-loop planning.

\bibliography{aaai2027}

@inproceedings{liu2023aerialvln,
  title={Aerialvln: Vision-and-language navigation for uavs},
  author={Liu, Shubo and Zhang, Hongsheng and Qi, Yuankai and Wang, Peng and Zhang, Yanning and Wu, Qi},
  booktitle={Proceedings of the IEEE/CVF International Conference on Computer Vision},
  pages={15384--15394},
  year={2023}
}

@inproceedings{lee2024citynav,
  title={Citynav: A large-scale dataset for real-world aerial navigation},
  author={Lee, Jungdae and Miyanishi, Taiki and Kurita, Shuhei and Sakamoto, Koya and Azuma, Daichi and Matsuo, Yutaka and Inoue, Nakamasa},
  booktitle={Proceedings of the IEEE/CVF International Conference on Computer Vision},
  pages={5912--5922},
  year={2025}
}

@article{xu2026vgas,
  title={VGAS: value-guided action-chunk selection for few-shot vision-language-action adaptation},
  author={Xu, Changhua and Yu, En and Xuan, Junyu and Lu, Jie},
  journal={arXiv preprint arXiv:2602.07399},
  year={2026}
}

@inproceedings{yu2026generalized,
author = {Yu, En and Lu, Jie and Zhang, Guangquan},
title = {Generalized Incremental Learning under Concept Drift across Evolving Data Streams},
year = {2026},
booktitle = {Proceedings of the ACM Web Conference 2026},
pages = {3905–3916},
numpages = {12},
series = {WWW '26},
doi = {10.1145/3774904.3792379}
}

@article{gao2024aerial,
  title={Aerial vision-and-language navigation via semantic-topo-metric representation guided LLM reasoning},
  author={Gao, Yunpeng and Wang, Zhigang and Jing, Linglin and Wang, Dong and Li, Xuelong and Zhao, Bin},
  journal={arXiv preprint arXiv:2410.08500},
  year={2024}
}

@article{zhao2025aerial,
  title={Aerial Vision-and-Language Navigation with Grid-based View Selection and Map Construction},
  author={Zhao, Ganlong and Li, Guanbin and Pan, Jia and Yu, Yizhou},
  journal={arXiv preprint arXiv:2503.11091},
  year={2025}
}

@inproceedings{wang2024towards,
  title={Towards realistic uav vision-language navigation: Platform, benchmark, and methodology},
  author={Wang, Xiangyu and Yang, Donglin and Kwan, Hohin and Chen, Jinyu and Li, Hongsheng and Liao, Yue and Liu, Si and others},
  booktitle={International Conference on Learning Representations},
  volume={2025},
  pages={7292--7310},
  year={2025}
}

@inproceedings{fan2023aerial,
  title={Aerial vision-and-dialog navigation},
  author={Fan, Yue and Chen, Winson and Jiang, Tongzhou and Zhou, Chun and Zhang, Yi and Wang, Xin},
  booktitle={Findings of the Association for Computational Linguistics: ACL 2023},
  pages={3043--3061},
  year={2023}
}

@article{xu2026aerialvla,
  title={AerialVLA: A Vision-Language-Action Model for UAV Navigation via Minimalist End-to-End Control},
  author={Xu, Peng and Deng, Zhengnan and Deng, Jiayan and Gu, Zonghua and Wan, Shaohua},
  journal={arXiv preprint arXiv:2603.14363},
  year={2026}
}

@article{zhang2025grounded,
  title={Grounded Vision-Language Navigation for UAVs with Open-Vocabulary Goal Understanding},
  author={Zhang, Yuhang and Yu, Haosheng and Xiao, Jiaping and Feroskhan, Mir},
  journal={arXiv preprint arXiv:2506.10756},
  year={2025}
}

@article{xu2025geonav,
  title={Geonav: Empowering mllms with explicit geospatial reasoning abilities for language-goal aerial navigation},
  author={Xu, Haotian and Hu, Yue and Gao, Chen and Zhu, Zhengqiu and Zhao, Yong and Li, Yong and Yin, Quanjun},
  journal={arXiv preprint arXiv:2504.09587},
  year={2025}
}

@inproceedings{ding2026history,
  title={History-enhanced two-stage transformer for aerial vision-and-language navigation},
  author={Ding, Xichen and Gao, Jianzhe and Pan, Cong and Wang, Wenguan and Qin, Jie},
  booktitle={Proceedings of the AAAI Conference on Artificial Intelligence},
  volume={40},
  number={22},
  pages={18225--18233},
  year={2026}
}

@inproceedings{fan2026htnav,
  title={HTNav: A Hybrid Navigation Framework with Tiered Structure for Urban Aerial Vision-and-Language Navigation},
  author={Fan, Chengjie and Pan, Cong and Liu, Zijian and Liu, Ningzhong and Qin, Jie},
  booktitle={Proceedings of the IEEE/CVF Conference on Computer Vision and Pattern Recognition},
  pages={10976--10985},
  year={2026}
}

@inproceedings{sautenkov2025uav,
  title={UAV-VLA: Vision-language-action system for large scale aerial mission generation},
  author={Sautenkov, Oleg and Yaqoot, Yasheerah and Lykov, Artem and Mustafa, Muhammad Ahsan and Tadevosyan, Grik and Akhmetkazy, Aibek and Cabrera, Miguel Altamirano and Martynov, Mikhail and Karaf, Sausar and Tsetserukou, Dzmitry},
  booktitle={2025 20th ACM/IEEE International Conference on Human-Robot Interaction (HRI)},
  pages={1588--1592},
  year={2025},
  organization={IEEE}

}

@inproceedings{sautenkov2025uavvlpa,
  title={Uav-vlpa*: Vision-language guided global-local uav mission planning from satellite imagery},
  author={Sautenkov, Oleg and Akhmetkazy, Aibek and Yaqoot, Yasheerah and Mustafa, Muhammad Ahsan and Tadevosyan, Grik and Lykov, Artem and Serpiva, Valerii and Tsetserukou, Dzmitry},
  booktitle={2025 IEEE International Conference on Robotics and Biomimetics (ROBIO)},
  pages={2354--2359},
  year={2025},
  organization={IEEE}
}

@inproceedings{su2025learning,
  title={Learning fine-grained alignment for aerial vision-dialog navigation},
  author={Su, Yifei and An, Dong and Chen, Kehan and Yu, Weichen and Ning, Baiyang and Ling, Yonggen and Huang, Yan and Wang, Liang},
  booktitle={Proceedings of the AAAI Conference on Artificial Intelligence},
  volume={39},
  number={7},
  pages={7060--7068},
  year={2025}
}

@inproceedings{chen2026aerialvla,
  title={AerialVLA: A Vision-Language-Action Model for Aerial Navigation with Online Dialogue},
  author={Chen, Jinyu and Li, Hongyu and Tang, Zongheng and Li, Xiaoduo and Wu, Wenjun and Liu, Si},
  booktitle={Proceedings of the AAAI Conference on Artificial Intelligence},
  volume={40},
  number={22},
  pages={18161--18169},
  year={2026}
}

@inproceedings{qi2026parse,
  title={Parse, Search, and Confirmation: Training-Free Aerial Vision-and-Dialog Navigation with Chain-of-Thought Reasoning and Structured Spatial Memory},
  author={Qi, Yu and Li, Hongyu and Huang, Shaofei and Hui, Tianrui and Wang, Yaxiong and Cheng, Lechao and Zhong, Zhun and Liu, Si and Wang, Meng},
  booktitle={Proceedings of the IEEE/CVF Conference on Computer Vision and Pattern Recognition},
  pages={23859--23868},
  year={2026}
}

@article{jiang2026spatialfly,
  title={SpatialFly: Geometry-Guided Representation Alignment for UAV Vision-and-Language Navigation in Urban Environments},
  author={Jiang, Wen and Huang, Kangyao and Wang, Li and Xu, Wang and Fan, Wei and Liu, Jinyuan and Liu, Shaoyu and Liang, Hanfang and Duan, Hongwei and Xu, Bin and others},
  journal={arXiv preprint arXiv:2603.21046},
  year={2026}
}

@article{lin2025openvln,
  title={OpenVLN: Open-world Aerial Vision-Language Navigation},
  author={Lin, Peican and Sun, Gan and Liu, Chenxi and Li, Fazeng and Ren, Weihong and Cong, Yang},
  journal={arXiv preprint arXiv:2511.06182},
  year={2025}
}

@article{jiang2026dynfly,
  title={DynFly: Dynamic-Aware Continuous Trajectory Generation for UAV Vision-Language Navigation in Urban Environments},
  author={Jiang, Wen and Liang, Hanfang and Wang, Li and Huang, Kangyao and Xu, Wang and Fan, Wei and Liu, Jinyuan and Liu, Shaoyu and Duan, Hongwei and Xu, Bin and others},
  journal={arXiv preprint arXiv:2606.31654},
  year={2026}
}

@article{xue2026see,
  title={See-and-Reach: Precise Vision-Language Navigation for UAVs within the Field of View},
  author={Xue, Fanfu and Yu, En and Shen, Yantian and Hu, Zhikun and Wang, Hongjun and Yang, Yang and Wang, Xindi and Sun, Jiande},
  journal={arXiv preprint arXiv:2606.20045},
  year={2026}
}

@inproceedings{zhang2025citynavagent,
  title={Citynavagent: Aerial vision-and-language navigation with hierarchical semantic planning and global memory},
  author={Zhang, Weichen and Gao, Chen and Yu, Shiquan and Peng, Ruiying and Zhao, Baining and Zhang, Qian and Cui, Jinqiang and Chen, Xinlei and Li, Yong},
  booktitle={Proceedings of the 63rd Annual Meeting of the Association for Computational Linguistics (Volume 1: Long Papers)},
  pages={31292--31309},
  year={2025}
}

@inproceedings{ning2026lookasidevln,
  title={LookasideVLN: direction-aware aerial vision-and-language navigation},
  author={Ning, Yuwei and Zhao, Ganlong and Qin, Yipeng and Liu, Si and Liu, Yang and Lin, Liang and Li, Guanbin},
  booktitle={Proceedings of the IEEE/CVF Conference on Computer Vision and Pattern Recognition},
  pages={32441--32450},
  year={2026}
}

@inproceedings{wang2026cotfly,
  title={CoTFly: Making UAVs Think Where to Fly Next Through Visual Chain-of-Thought Reasoning},
  author={Wang, Meiqi and Xu, Longnyu and Liu, Jun and Li, Hewu and Qiu, Han},
  booktitle={Proceedings of the IEEE/CVF Conference on Computer Vision and Pattern Recognition},
  pages={1482--1491},
  year={2026}
}

@article{zheng2026think,
  title={Think Like a Pilot: Fine-Grained Long-Horizon UAV Navigation},
  author={Zheng, Xiangyi and Wang, Xiangyu and Liao, Qinan and Tang, Zimu and Liao, Yue and Lyu, Dongyue and Wang, Guodong and Liu, Junjie and Liu, Si},
  journal={arXiv preprint arXiv:2606.06836},
  year={2026}
}

@inproceedings{wu2025aeroduo,
  title={AeroDuo: Aerial Duo for UAV-based Vision and Language Navigation},
  author={Wu, Ruipu and Zhang, Yige and Chen, Jinyu and Huang, Linjiang and Zhang, Shifeng and Zhou, Xu and Wang, Liang and Liu, Si},
  booktitle={Proceedings of the 33rd ACM International Conference on Multimedia},
  pages={2576--2585},
  year={2025}
}

@article{wang2026uav,
  title={Uav-flow colosseo: A real-world benchmark for flying-on-a-word uav imitation learning},
  author={Wang, Xiangyu and Yang, Donglin and Liao, Yue and Zheng, Wenhao and Dai, Bin and Li, Hongsheng and Liu, Si and others},
  journal={Advances in Neural Information Processing Systems},
  volume={38},
  year={2026}
}

@article{bai2025qwen3,
  title={Qwen3-vl technical report},
  author={Bai, Shuai and Cai, Yuxuan and Chen, Ruizhe and Chen, Keqin and Chen, Xionghui and Cheng, Zesen and Deng, Lianghao and Ding, Wei and Gao, Chang and Ge, Chunjiang and others},
  journal={arXiv preprint arXiv:2511.21631},
  year={2025}
}

@misc{bai2025qwen25vltechnicalreport,
      title={Qwen2.5-VL Technical Report}, 
      author={Shuai Bai and Keqin Chen and Xuejing Liu and Jialin Wang and Wenbin Ge and Sibo Song and Kai Dang and Peng Wang and Shijie Wang and Jun Tang and Humen Zhong and Yuanzhi Zhu and Mingkun Yang and Zhaohai Li and Jianqiang Wan and Pengfei Wang and Wei Ding and Zheren Fu and Yiheng Xu and Jiabo Ye and Xi Zhang and Tianbao Xie and Zesen Cheng and Hang Zhang and Zhibo Yang and Haiyang Xu and Junyang Lin},
      year={2025},
      eprint={2502.13923},
      archivePrefix={arXiv},
      primaryClass={cs.CV},
      url={https://arxiv.org/abs/2502.13923}, 
}

@article{gao2025openfly,
  title={OpenFly: A versatile toolchain and large-scale benchmark for aerial vision-language navigation},
  author={Gao, Yunpeng and Li, Chenhui and You, Zhongrui and Liu, Junli and Li, Zhen and Chen, Pengan and Chen, Qizhi and Tang, Zhonghan and Wang, Liansheng and Yang, Penghui and others},
  journal={arXiv e-prints},
  pages={arXiv--2502},
  year={2025}
}

@article{hu2022sensaturban,
  title={Sensaturban: Learning semantics from urban-scale photogrammetric point clouds},
  author={Hu, Qingyong and Yang, Bo and Khalid, Sheikh and Xiao, Wen and Trigoni, Niki and Markham, Andrew},
  journal={International Journal of Computer Vision},
  volume={130},
  number={2},
  pages={316--343},
  year={2022},
  publisher={Springer}
}

@inproceedings{xiao2025uav,
  title={Uav-on: A benchmark for open-world object goal navigation with aerial agents},
  author={Xiao, Jianqiang and Sun, Yuexuan and Shao, Yixin and Gan, Boxi and Liu, Rongqiang and Wu, Yanjin and Guan, Weili and Deng, Xiang},
  booktitle={Proceedings of the 33rd ACM International Conference on Multimedia},
  pages={13023--13029},
  year={2025}
}

@ARTICLE{10938647,
  author={Lin, Bingqian and Nie, Yunshuang and Wei, Ziming and Chen, Jiaqi and Ma, Shikui and Han, Jianhua and Xu, Hang and Chang, Xiaojun and Liang, Xiaodan},
  journal={IEEE Transactions on Pattern Analysis and Machine Intelligence}, 
  title={NavCoT: Boosting LLM-Based Vision-and-Language Navigation via Learning Disentangled Reasoning}, 
  year={2025},
  volume={47},
  number={7},
  pages={5945-5957},
  }

@inproceedings{
hu2022lora,
title={Lora: Low-rank adaptation of large language models.},
author={Edward J Hu and Yelong Shen and Phillip Wallis and Zeyuan Allen-Zhu and Yuanzhi Li and Shean Wang and Lu Wang and Weizhu Chen},
booktitle={International Conference on Learning Representations},
year={2022}
}

@article{zheng2026onfly,
  title={OnFly: Onboard Zero-Shot Aerial Vision-Language Navigation toward Safety and Efficiency},
  author={Zheng, Guiyong and Ban, Yueting and Zhang, Mingjie and Zheng, Juepeng and Zhou, Boyu},
  journal={arXiv preprint arXiv:2603.10682},
  year={2026}
}

@article{wu2025vla,
  title={VLA-AN: An Efficient and Onboard Vision-Language-Action Framework for Aerial Navigation in Complex Environments},
  author={Wu, Yuze and Zhu, Mo and Li, Xingxing and Du, Yuheng and Fan, Yuxin and Li, Wenjun and Han, Zhichao and Zhou, Xin and Gao, Fei},
  journal={arXiv preprint arXiv:2512.15258},
  year={2025}
}

@article{zhang2026apex,
  title={APEX: A Decoupled Memory-based Explorer for Asynchronous Aerial Object Goal Navigation},
  author={Zhang, Daoxuan and Chen, Ping and Xia, Xiaobo and Su, Xiu and Zhen, Ruichen and Xiao, Jianqiang and Yang, Shuo},
  journal={arXiv preprint arXiv:2602.00551},
  year={2026}
}

@article{liu2026imagineuav,
  title={ImagineUAV: Aerial Vision-Language Navigation via World-Action Modeling and Kinodynamic Planning},
  author={Liu, Xuchen and Huang, Jiawei and Xia, Shihao and Liu, Bingxi and Cui, Jinqiang and Yang, Jiankun},
  journal={arXiv preprint arXiv:2606.01205},
  year={2026}
}

@article{zheng2026worldfly,
  title={WorldFly: A World-Model-Based Vision-Language-Action Model for UAV Navigation},
  author={Zheng, Shengtao and Li, Kai and Zhang, Weichen and Meng, Yu and Gao, Chen and Chen, Xinlei and Li, Yong and Zhang, Xiao-Ping},
  journal={arXiv preprint arXiv:2606.06147},
  year={2026}
}

@article{sun2026autofly,
  title={Autofly: Vision-language-action model for UAV autonomous navigation in the wild},
  author={Sun, Xiaolou and Si, Wufei and Ni, Wenhui and Li, Yuntian and Wu, Dongming and Xie, Fei and Guan, Runwei and Xu, He-Yang and Ding, Henghui and Wu, Yuan and others},
  journal={arXiv preprint arXiv:2602.09657},
  year={2026}
}

@article{foead2021systematic,
  title={A systematic literature review of A* pathfinding},
  author={Foead, Daniel and Ghifari, Alifio and Kusuma, Marchel Budi and Hanafiah, Novita and Gunawan, Eric},
  journal={Procedia Computer Science},
  volume={179},
  pages={507--514},
  year={2021},
  publisher={Elsevier}
}

@inproceedings{zhou2026memory,
  title={Memory-Augmented Scene Understanding and Exploration for Open-World Aerial Object-Goal Navigation},
  author={Zhou, Jiacong and Miao, Jiaxu and Lin, Yourun and Wang, Xianyun and Xiao, Jun and Yu, Jun},
  booktitle={Proceedings of the IEEE/CVF Conference on Computer Vision and Pattern Recognition},
  pages={21616--21626},
  year={2026}
}

@article{liu2024navagent,
  title={Navagent: Multi-scale urban street view fusion for uav embodied vision-and-language navigation},
  author={Liu, Youzhi and Yao, Fanglong and Yue, Yuanchang and Xu, Guangluan and Sun, Xian and Fu, Kun},
  journal={arXiv preprint arXiv:2411.08579},
  year={2024}
}

\clearpage
\newpage
\appendix
\section*{Appendix}

\setcounter{secnumdepth}{2}

\section{Experiment Details}
\label{app:experiment_details}

\subsection{Implementation and Evaluation Settings}
\label{app:parameter_settings}
DBFly is fine-tuned on the training set of UAV-VLN-FOV~\cite{xue2026see} and evaluated in a closed-loop manner within simulation environments built using Unreal Engine and AirSim. A fixed parameter configuration is used for DBFly across all experiments. The forward-facing camera in the simulation environments has a horizontal field of view of $\pi$ radians. Accordingly, $\alpha=\pi/2$ defines the outer angular bound of the front-view region, and $\gamma=\pi/6$ separates the central \texttt{front} region from the lateral \texttt{front\_left} and \texttt{front\_right} regions. The threshold parameters are set to $\epsilon_{\kappa}=0$, $\epsilon_l=\epsilon_{\mu z}=\epsilon_r=0.25$, $\epsilon_d=\epsilon_f=\epsilon_m=\epsilon_c=\epsilon_{\mu x}=0.5$, and $\epsilon_s=2$. The short-horizon waypoint sequence length is fixed at $K=5$. All angular quantities are expressed in radians, whereas quantities describing displacement and motion are expressed in meters. Each reported test result is based on a single evaluation run under the corresponding experimental setting.

\subsection{Real-World Deployment}
\begin{figure}
\centering
\includegraphics[width=0.4\textwidth]{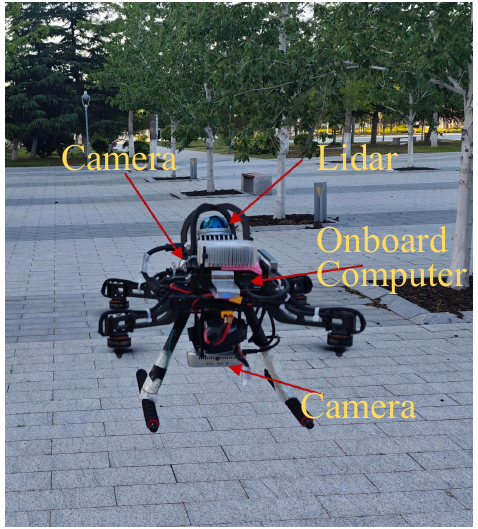}
\caption{UAV platform for real-world flight experiments.}
\label{fig:fig_uav}
\end{figure}

The real-world UAV platform is shown in Fig.~\ref{fig:fig_uav}. The UAV is equipped with a Livox Mid-360 LiDAR for environmental mapping and localization. Two Intel RealSense D435i cameras are mounted in forward-facing and downward-facing orientations to capture visual observations at a resolution of $1280\times720$. An NVIDIA Jetson Orin NX with 16 GB of memory serves as the onboard computer, while DBFly is deployed on a remote computing platform equipped with an NVIDIA RTX 4090 GPU for model inference. The onboard computer and the remote computing platform communicate over a dedicated wireless network. During flight, the onboard computer receives the language instruction and captures egocentric visual observations, which are jointly transmitted to the remote computing platform for model inference. The predicted navigation decision and future waypoints are then returned to the onboard computer, where they are converted into flight-controller-compatible setpoints and sent to the flight controller for execution.

\begin{table*}
\centering
\begin{tabular}{ccccccccccccc} 
  \toprule
  \multirow{2}{*}{\textbf{Strategies}} & \multicolumn{4}{c}{Test} & \multicolumn{4}{c}{Test UO}  &  \multicolumn{4}{c}{Test US}\\
  \cmidrule(lr){2-5}   \cmidrule(lr){6-9}   \cmidrule(lr){10-13} 
  &SR \textuparrow & OSR \textuparrow &NE \textdownarrow & SPL \textuparrow &SR \textuparrow & OSR \textuparrow &NE \textdownarrow & SPL \textuparrow &SR \textuparrow & OSR \textuparrow &NE \textdownarrow & SPL \textuparrow  \\
  \midrule
w/o C\&V  & 30.92 & 53.29 & 30.07 & 21.33 & 27.44 & 36.59 & 35.75 & 21.27 & 30.64 & 41.04 & 37.08 & 19.10 \\
w/o C\&V\&S  & 43.42 & 59.51 & 28.41 & 30.83 & 30.49 & 42.07 & 32.21 & 22.63 & 32.37 & 44.51 & 35.15 & 20.25 \\
w/o S & 32.89 & 55.26 & 28.98 & 21.90 & 21.34 & 40.24 & 34.89 & 12.95 & 22.54 & 37.57 & 36.80 & 12.89 \\
\midrule
DBFly & \textbf{62.50} & \textbf{71.05} & \textbf{16.83} & \textbf{42.37}  & \textbf{51.83} & \textbf{63.41} & \textbf{21.39} & \textbf{36.07} &\textbf{49.13} & \textbf{57.23} & \textbf{22.21} & \textbf{31.24} \\
    \midrule
\end{tabular}
\caption{Ablation study of the sampling strategy. w/o denotes removal. w/o C\&V removes the implicit flight corridor and vision-guided spatial deliberation; w/o C\&V\&S further removes sampling; w/o S removes sampling from the full DBFly.}
\label{app_ablation_studies}
\end{table*}

\subsection{Complete Prompt and Output Example}
\label{app:complete_prompt}
This section presents a complete example of the fixed system prompt, per-step user prompt, and structured model response used by DBFly. In the user prompt, only the \textit{Current Corridor State} is updated at each navigation step, while all other content remains unchanged. The two egocentric visual observations are represented by image placeholders. We use the language instructions provided in~\cite{xue2026see}, each of which contains a target-reaching instruction and a coarse directional cue. We retain the target-reaching component as the navigation instruction and extract the directional cue as the initial direction prior. For example, the original annotation "Fly to the white truck on the city street. It's down below you." is converted into "Instruction: Fly to the white truck on the city street" and "Initial direction prior: \texttt{down}". The complete prompt and corresponding output example are presented below.

\begingroup

\setlength{\fboxsep}{4pt}
\setlength{\fboxrule}{0.45pt}
\setlength{\parindent}{0pt}
\emergencystretch=1em

\newcommand{\promptbox}[2]{%
\noindent
\fbox{%
\begin{minipage}{\dimexpr\columnwidth-2\fboxsep-2\fboxrule\relax}
\footnotesize
\raggedright
\setlength{\parindent}{0pt}
\setlength{\parskip}{0pt}

\textbf{#1}

\vspace{2pt}
\hrule
\vspace{4pt}

#2

\end{minipage}%
}
\par\vspace{4pt}
}

\promptbox{System Prompt (Fixed)}{%
You are an intelligent UAV navigation agent with spatial geometric
reasoning ability. You will be given two images in fixed order:
(1) Current Front View and (2) Current Downward View. The UAV body
frame is defined as $x$ forward, $y$ right, and $z$ down. The current
corridor state is computed from UAV odometry and should be respected
as a soft geometric condition. You should generate a
corridor-consistent navigation decision and five future waypoints in
the current UAV body frame.
}

\promptbox{User Prompt (Per Step): Inputs and Required Output}{%
\texttt{<image>}

\texttt{<image>}

\vspace{3pt}

\textbf{Instruction:}
Fly to the white truck on the city street.

\vspace{2pt}

\textbf{Initial direction prior:}
\texttt{down}.

\vspace{2pt}

\textbf{Current corridor state:}
\texttt{corridor\_down\_approach}.

\vspace{3pt}

Return a JSON object with keys:
\texttt{"target\_direction"},
\texttt{"diagnosis"},
\texttt{"action"},
\texttt{"stop"}, and
\texttt{"waypoints\_body"}.
}

\promptbox{User Prompt (Per Step): Output Constraints}{%
\smallskip
Constraints:

\textbf{1.} \texttt{target\_direction} describes the target's current coarse region relative to the UAV body frame. It must be one of: \texttt{front\_left}, \texttt{front}, \texttt{front\_right}, \texttt{down}.

\smallskip

\textbf{2.} \texttt{current\_corridor\_state} is an input condition and is one of: \texttt{corridor\_centered}, \texttt{corridor\_left\_deviation}, \texttt{corridor\_right\_deviation}, \texttt{corridor\_down\_approach}.

\smallskip

\textbf{3.} \texttt{diagnosis} must be one of: \texttt{forward\_progress}, \texttt{lateral\_realign\_left}, \texttt{lateral\_realign\_right}, \texttt{vertical\_realign}, \texttt{approach\_and\_contract}, \texttt{terminal\_converging}, \texttt{fine\_realign}.
}

\promptbox{User Prompt (Per Step): Output Constraints}{%
\smallskip

\textbf{4.} \texttt{action} must be one of: \texttt{forward}, \texttt{forward\_left}, \texttt{forward\_right}, \texttt{forward\_down}, \texttt{descend}, \texttt{align\_left\_down}, \texttt{align\_right\_down}, \texttt{hover}.

\smallskip

\textbf{5.} \texttt{action} must be geometrically consistent with \texttt{current\_corridor\_state}, \texttt{target\_direction}, and \texttt{waypoints\_body}.

\smallskip

\textbf{6.} \texttt{stop} must be \texttt{yes} only when the UAV is close to the target and the motion is stably converging; otherwise \texttt{no}.

\smallskip

\textbf{7.} \texttt{waypoints\_body} must contain exactly 5 future waypoints in the current UAV body frame.

\smallskip

\textbf{8.} Output valid JSON only.
}

\promptbox{Model Response (Example)}{%
{\ttfamily
\{\\
\hspace*{1em}"target\_direction": "down",\\
\hspace*{1em}"diagnosis": "vertical\_realign",\\
\hspace*{1em}"action": "descend",\\
\hspace*{1em}"stop": "no",\\
\hspace*{1em}"waypoints\_body": [\\
\hspace*{2em}[0.0645, 0.0053, 0.9899],\\
\hspace*{2em}[0.1435, 0.0016, 2.0044],\\
\hspace*{2em}[0.2030, -0.0026, 2.9681],\\
\hspace*{2em}[0.2696, 0.0026, 3.9820],\\
\hspace*{2em}[0.3547, 0.0033, 4.9962]\\
\hspace*{1em}]\\
\}
}
}

\endgroup

\section{Additional Experiments}
\label{app:additional_experiments}
\subsection{Effect of Sampling} 
Table~\ref{app_ablation_studies} disentangles the effect of sampling from that of the proposed spatial deliberation framework. When both the implicit flight corridor and vision-guided spatial deliberation are removed, retaining sampling fails to improve performance. Specifically, \textit{w/o C\&V} performs worse than \textit{w/o C\&V\&S}, with SR decreasing by 12.50, 3.05, and 1.73 percentage points on Test, Test UO, and Test US, respectively. This result indicates that merely rebalancing the action distribution through sampling is insufficient for the model to acquire the corresponding spatial maneuvering capability and may even degrade performance in the absence of explicit spatial deliberation. In contrast, removing sampling from the full DBFly model substantially reduces SR by 29.61, 30.49, and 26.59 percentage points and SPL by 20.47, 23.12, and 18.35 percentage points across the three test sets. These results demonstrate a strong interaction between sampling and spatial deliberation: sampling becomes effective only when it rebalances and reinforces the structured spatial decisions explicitly learned by DBFly. Therefore, DBFly's performance gains do not stem from sampling itself, but primarily from the proposed spatial deliberation framework, with sampling serving as a complementary training strategy.

\subsection{Qualitative Results}
\begin{figure*}
\centering
\includegraphics[width=\textwidth]{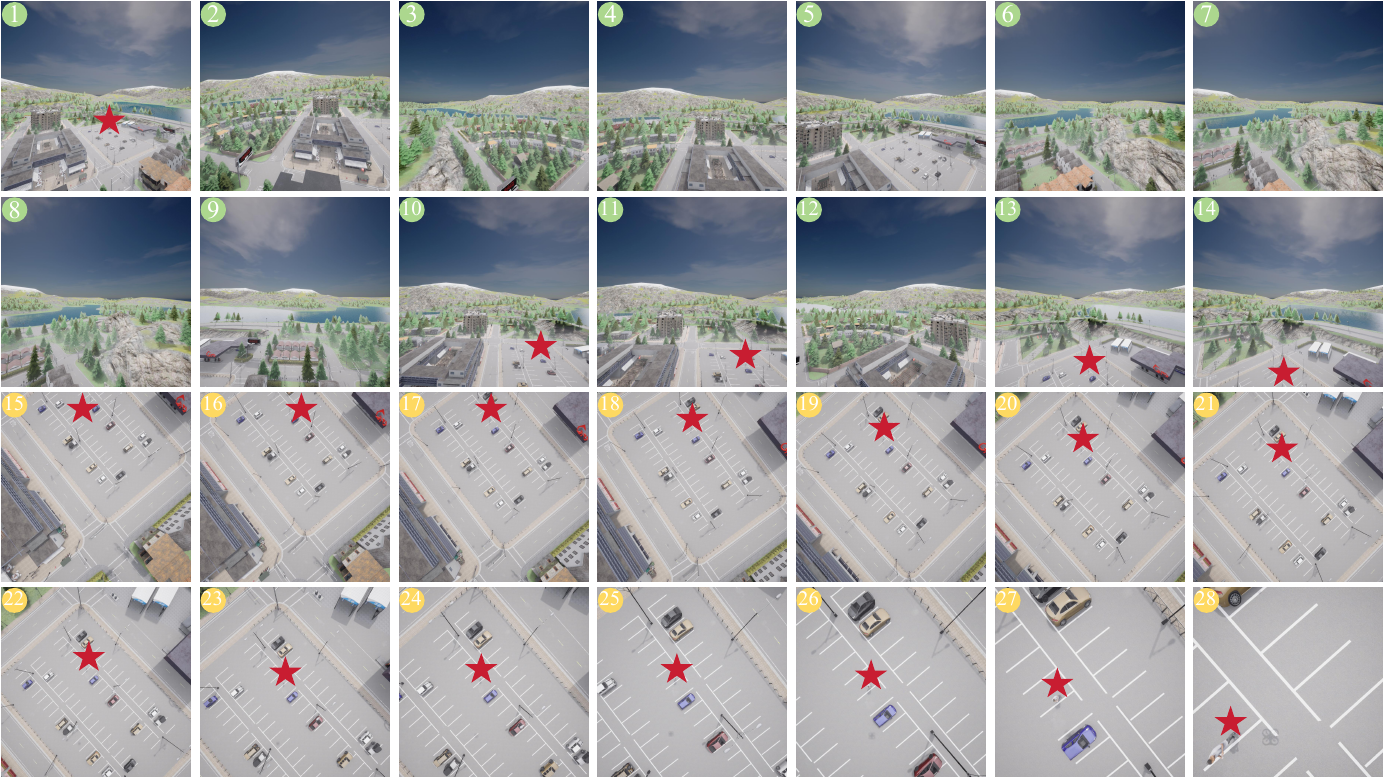}
\caption{Visualization of DBFly navigation in the high-fidelity simulation environment. Instruction: Fly to the man in the parking area. Initial direction prior: front. The red star indicates the approximate target location, while its absence means that the target is outside the UAV's current field of view. Numbers highlighted in green denote front-view images, while those highlighted in yellow denote downward-view images.}
\label{fig_app_simulation_1}
\end{figure*}

\begin{figure*}
\centering
\includegraphics[width=\textwidth]{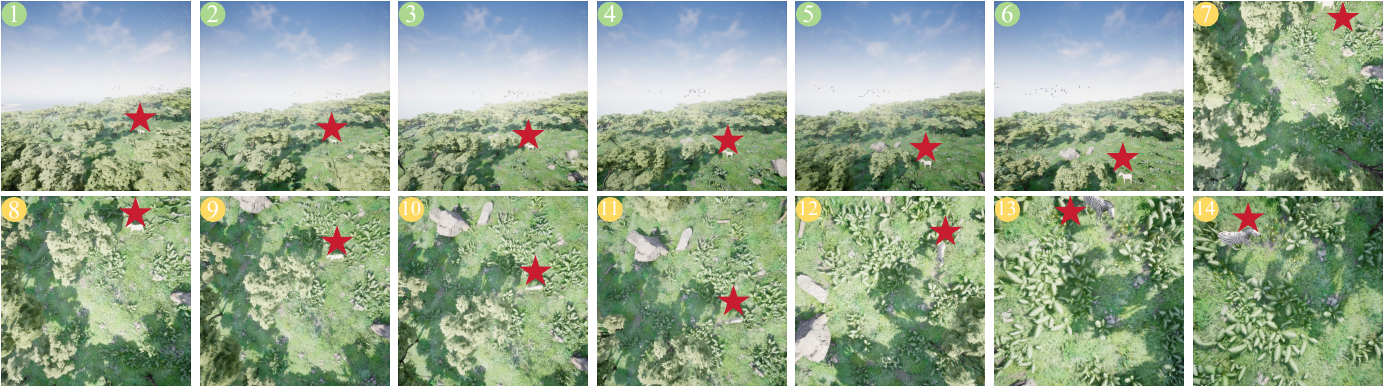}
\caption{Visualization of DBFly navigation in the high-fidelity simulation environment. Instruction: Fly to the zebra in the grassy area with scattered boulders. Initial direction prior: front\_right. The same annotation convention as that in Fig.~\ref{fig_app_simulation_1} is adopted.}
\label{fig_app_simulation_2}
\end{figure*}

\begin{figure*}
\centering
\includegraphics[width=\textwidth]{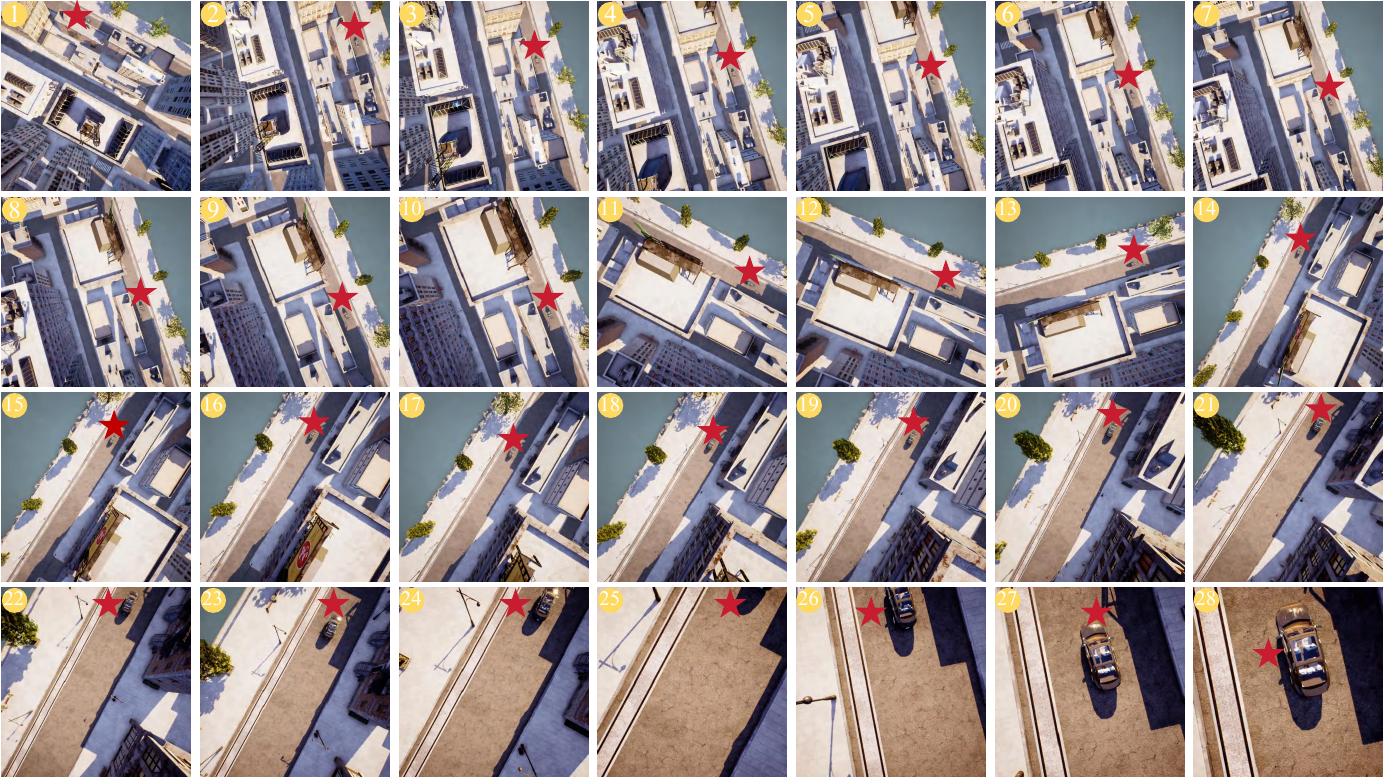}
\caption{Visualization of DBFly navigation in the high-fidelity simulation environment. Instruction: Fly to the black car parked on the street near the brick buildings. Initial direction prior: down. The same annotation convention as that in Fig.~\ref{fig_app_simulation_1} is adopted.}
\label{fig_app_simulation_3}
\end{figure*}

\begin{figure*}
\centering
\includegraphics[width=\textwidth]{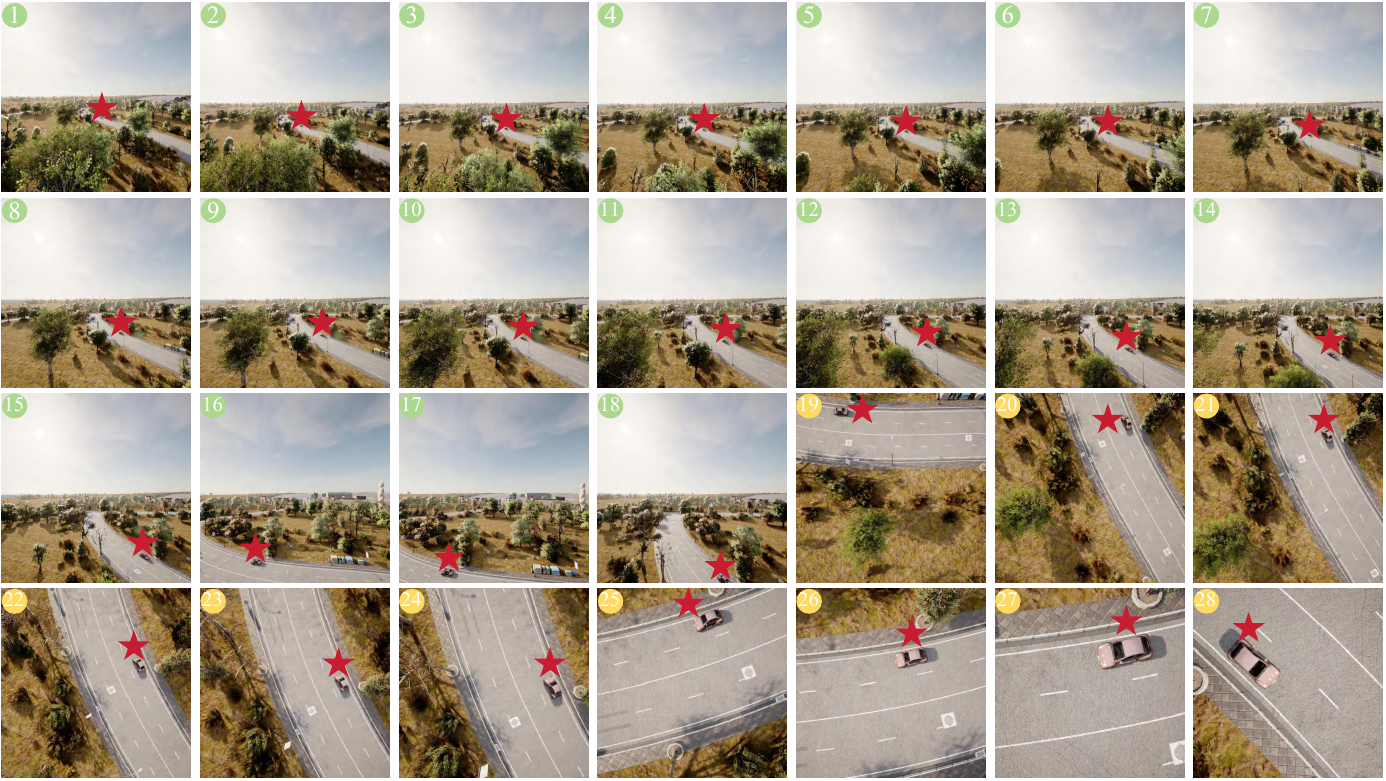}
\caption{Visualization of DBFly navigation in the high-fidelity simulation environment. Instruction: Fly to the brown car on the gray road with white lane markings. Initial direction prior: front. The same annotation convention as that in Fig.~\ref{fig_app_simulation_1} is adopted.}
\label{fig_app_simulation_4}
\end{figure*}

We further visualize the UAV observations collected throughout multiple navigation episodes, as shown in Figs.~\ref{fig_app_simulation_1}--\ref{fig_app_simulation_4}. Fig.~\ref{fig_app_simulation_1} shows that, when the target appears too small and blurred in the initial observation to be reliably recognized, DBFly autonomously explores the surrounding environment. Once the target is identified and anchored, DBFly progressively approaches it and ultimately comes to a stable stop in its vicinity. Fig.~\ref{fig_app_simulation_2} demonstrates that DBFly can reliably identify and accurately approach a target even when it lies outside the \texttt{front} region. Fig.~\ref{fig_app_simulation_3} illustrates that, during high-altitude descent, DBFly maintains lateral stability while progressively aligning with and descending toward the target. Fig.~\ref{fig_app_simulation_4} further demonstrates DBFly's ability to maintain continuous target anchoring throughout the flight. These qualitative results further validate DBFly's robust spatial maneuvering and reliable stopping capabilities.

\end{document}